\documentclass[12pt]{article}

\def\modifymargins#1#2{
\newdimen\addtoh
\newdimen\addtow
\addtoh=#1
\addtow=#2

\advance\topmargin by -\addtoh
\multiply\addtoh by 2
\advance\textheight by \addtoh

\advance\oddsidemargin by -\addtow
\advance\evensidemargin by -\addtow
\multiply\addtow by 2
\advance\textwidth by \addtow
}

\makeatletter

\long\def\state#1#2#3#4#5{%
\def\statelabel{\label{#1}}
\def\st@atehere{#3#4}
\ifx\st@atehere\empty\expandafter\def\csname st@atementempty#1\endcsname{Y}\else#3\fi#4%
\expandafter\ifx\csname stateonce\endcsname\relax\relax\else #5\fi
\ifx#2\relax\else
\expandafter\def\csname st@atementlabel#1\endcsname{#2}
\fi
\expandafter\def\csname st@atement#1\endcsname{#4#5}
}

\newcount\restatec@unter

\def\restate#1{
\expandafter\ifx\csname stateonce\endcsname\relax
\expandafter\let\expandafter\what\csname st@atementlabel#1\endcsname
\expandafter\ifx\what\relax
\csname st@atement#1\endcsname
\else
\def\rest@teempty{Y}
\expandafter\ifx\csname st@atementempty#1\endcsname\rest@teempty
\def\statelabel{\label{#1}}
\csname st@atement#1\endcsname
\else
\expandafter\let\expandafter\counter\csname c@\what\endcsname
\restatec@unter=\counter
\counter=\expandafter\ifx\csname r@#1\endcsname\relax0\else\@kernel@ref{#1}\fi
\advance\counter by -1
\let\statelabel=\relax
\csname st@atement#1\endcsname
\counter=\restatec@unter
\fi
\fi
\fi
}

\makeatother

\def\skipproof#1{\def\nope{\iffalse}\def\go{\iftrue}%
\ifx#1y[proof commented out]\let\next=\nope\else\let\next=\go\fi\next}

\def\draft{\iftrue\begingroup\par\em=============[draft]=============\par}
\def\enddraft{\par=============[/draft]=============\par\endgroup\fi}
\def\draft{\iffalse}
\let\enddraft=\fi

\long\def\nop#1{}
\def\p{{\rm P}}
\def\np{{\rm NP}}
\def\conp{{\rm coNP}}
\def\D#1{\mbox{$\Delta^p_{#1}$}}

\def\true{{\sf true}}
\def\false{{\sf false}}
\def\proof{\noindent {\sl Proof.\ \ }}
\def\qed{\hfill{\boxit{}}
  \ifdim\lastskip<\medskipamount \removelastskip\penalty55\medskip\fi}
\long\def\boxit#1{\vbox{\hrule\hbox{\vrule\kern3pt
                  \vbox{\kern3pt#1\kern3pt}\kern3pt\vrule}\hrule}}

\def\color[#1]#2{}
\def\possnewtheorem#1#2{
\expandafter\ifx\csname #1\endcsname\relax
\newtheorem{#1}{#2}
\fi
}

\possnewtheorem{theorem}{Theorem}
\possnewtheorem{corollary}{Corollary}
\possnewtheorem{lemma}{Lemma}
\possnewtheorem{definition}{Definition}
\possnewtheorem{algorithm}{Algorithm}
\possnewtheorem{counterexample}{Counterexample}

\title{Can we forget how we learned?\\
Doxastic redundancy in iterated belief revision}
\author{Paolo Liberatore%
\thanks{
DIAG, Sapienza University of Rome,
{\tt liberato@diag.uniroma1.it}
}
}

\begin{document}

\maketitle

\draft

FARE: separator, newpage, draft, FARE

\enddraft

\draft

sviluppi ulteriori:

FARE: anche per gli OCF esiste una rappresentazione sintattica, un insieme di
(formula,numero) invece di (mondo,numero); il punto non e' tanto la esistenza
di una rappresentazione sintattica, ma quanto la sua compattezza, che poi e'
quello che indica se e' realistica

questo articolo riassume i meccanismi: paper-5.pdf
InfOCF-Lib: A Java Library for OCF-based Conditional Inference,
Steven Kutsch

"Several semantics have been proposed for defining nonmonotonic inference
relations over sets of such rules. Examples are Lewis' system of spheres [17],
conditional objects evaluated using Boolean intervals [9], or possibility
measures [8, 10]."

"Here, we will consider Spohn's ranking functions [20] that assign degrees of
disbelief to propositional interpretations. C-Representations [13, 14] are a
subset of all ranking functions for a knowledge base, which can conveniently be
calculated by solving a constraint satisfaction problem dependent on the
knowledge base. [...] the tool InfOCF [...] allows the user to load knowledge
bases, calculate admissible ranking functions (in particular c-representations
and system Z [19]), and perform inference using these sets of ranking
functions. System P entailment [1] was also implemented."

"several new theoretical approaches have been proposed such as different
notions of minimal c-representations [2] and inference usinsets of ranking
functions [5, 7]."

FARE: altri sistemi pratici, rivedere cosa fanno di preciso:

1812.08313.pdf
Iterated belief revision under resource constraints: logic as geometry
Dan P. Guralnik, Daniel E. Koditschek

aaai.v33i01.33013076.pdf
Iterated Belief Base Revision: A Dynamic Epistemic Logic Approach
Marlo Souza, Alvaro Moreira, Renata Vieira
sembra includere lexicographic o essere equivalente

sistemi che estendono lexicographic:

- priority graph
  aaai.v33i01.33013076.pdf
  Iterated Belief Base Revision: A Dynamic Epistemic Logic Approach
  Marlo Souza, Alvaro Moreira, Renata Vieira
  2019

- combinazioni lessicografiche di ordinamenti generici
  document (11).pdf
  Operators and Laws for Combining Preference Relations
  Hajnal Andreka, Mark Ryan, Pierre-Yves Schobbens
  2002

FARE: un'altra direzione futura e' misurare quanta informazione si perde con il
forget, invece di vedere solo se la perdita e' zero; e' analogo a misurare il
grado di inconsistenza, quindi non e' per niente impossibile

FARE: vedere altro in iterated-revision.txt

\enddraft

\begin{abstract}

Forgetting a belief acquisition episode may not cause information loss because
of the others. Checking whether it does is not obvious, as the contribution of
each belief revision is not isolated from the others, and the same information
may be given not directly but by deduction. An algorithm for checking whether
forgetting reduces information is given for a number of iterated belief
revision operators: lexicographic, natural, severe, plain severe, moderate
severe, restrained, very radical and full meet revisions. It may take
exponential time in the worst case, which is expected given that the problem is
\conp-hard, even in the Horn restriction. It is in \conp\  for homogeneous
sequences of lexicographic revisions.

\end{abstract}

\subsection*{Declarations}

\noindent
{\bf Funding:} this article is supported by PRIN project ``From 2D to 3D,
Surgical and legal medical aspects'' and by the SAPIENZA project ``Riduzione di
formule logiche''.

\noindent
{\bf Conflicts of interest/Competing interests:}
the author has no conflicts of interest to declare that are relevant to the content of this article

\noindent
{\bf Availability of data and material (data transparency):}
no data is used in this article

\noindent
{\bf Code availability (software application or custom code):}
code is freely available online at {\tt https://github.com/paololiberatore}

\noindent
{\bf Authors' contributions:}
single author did it all.

\section{Introduction}
\label{introduction}

People forget. Computers delete. Paper burns, parchments are washed and
rewritten~\cite{gipp-etal-25}, stones broken and reused~\cite{gonn-10}, oral
accounts no longer told~\cite{klei-89}.

Forgetting happens for a number of reasons:
{} psychology~\cite{myer-04},
{} politics~\cite{conn-17},
{} privacy~\cite{gonc-etal-17},
{} memory bounds~\cite{eite-kern-19},
{} reasoning simplification~\cite{delg-wang-15,erde-ferr-07,wang-etal-05},
{} knowledge clarification~\cite{delg-17},
{} knowledge limitations~\cite{fagi-etal-95,raja-etal-14},
{} searching reduction~\cite{smyt-kean-95},
{} information merging~\cite{wang-etal-14},
{} consistency restoration~\cite{lang-marq-10}.
Yet, information may not be lost. People may remember similar things, files
may be duplicated, documents copied.

Forgetting may not reduce information thanks to the unforgotten. Someone may
not remember the individual steps for dividing thirty-four by twelve, but still
reconstruct them from the general principles of operating by columns.

\draft
\bf acknowledge referee for the example
\enddraft

The belief revision process is a continous acquisition and forgetting of
beliefs. Logical formalizations~\cite{spoh-88,ryan-91,naya-94,bout-96-a,%
darw-pear-97,libe-97-c,benf-etal-99,benf-etal-00,koni-pere-00,roor-etal-03,%
zhan-04,hunt-delg-07,boot-etal-06,boot-nitt-08,rott-09,rott-12,libe-15,%
schw-etal-22,libe-23,libe-24} tell how to acquire---not how to forget. This is
the focus of this article: forgetting. Does forgetting a belief revision
episode cause information loss? Other beliefs may supply the same information.
The exact same, or related information from which it can be deduced.

The frame of iterated belief revision hinges on the current beliefs, ordered by
their strength. Every belief acquisition is a logical formula. They form a
sequence, a history of revisions that change the belief strength order, the
doxastic state. Does it change the same if some revision episodes are
forgotten? If it does, forgetting is harmless---no information is lost. Or
useless---information is not erased.

Information reduction may be the very aim of forgetting, like in
politics~\cite{conn-17} or privacy~\cite{gonc-etal-17}. It may be unintended,
like in psychology~\cite{norb-15} or knowledge clarification~\cite{delg-17}. It
may be unwanted, like in reasoning
simplification~\cite{delg-wang-15,erde-ferr-07,wang-etal-05}, searching
reduction~\cite{smyt-kean-95}, information merging~\cite{wang-etal-14} or
consistency restoration~\cite{lang-marq-10}. Whatever the reason for
forgetting, information reduction may result. Or not.

The algorithm in Section~\ref{algorithm} tells whether it does. It checks the
effect of removing revisions from a sequence of lexicographic revisions. It
establishes whether the doxastic state changes or not. Whether information is
lost. Intended or not, wanted or not, this may result. Aim or drawback or side
effect, it is a foremost outcome of forgetting.

The algorithm core establishes whether two sequences of
revisions~\cite{spoh-88,naya-94} have the same effect. It is limited to
lexicographic revisions (episodes of acquisition of information expressed by a
propositional formula) starting from the tabula-rasa doxastic state (the state
of not having any belief). The apparent limitations of the initial state and
kind of revision are overcome by adapting an arbitrary doxastic state and
sequence of revisions of possibly heterogeneous other kinds, like
natural~\cite{bout-96-a}, restrained~\cite{boot-meye-06} or
severe~\cite{rott-06}: every doxastic state is the outcome of a sequence of
lexicographic revisions~\cite{libe-24}; every sequence of these revisions
translates into a lexicographic sequence~\cite{libe-23}.

The algorithm may take exponential running time. This is unsurprising since
comparing the effects of two arbitrary sequences of revisions is \conp-complete
both for arbitrary propositional sequences and for Horn formulae. The proofs
extend to the specific case of removing a revision from a sequence, the
relevant subcase of forgetting a revision episode.

After the formal definitions are given in 
Section~\ref{definitions}, the results of this article are proven:

\begin{itemize}

\item an algorithm for establishing the equivalence between two sequences of
revisions, in Section~\ref{algorithm};

\item a necessary and sufficient condition to the redundancy of a lexicographic
revision, in Section~\ref{condition};

\item the equivalence and redundancy problems are proved \conp-complete for
arbitrary propositional formulae, in Section~\ref{general};

\item they are also \conp-complete in the Horn restriction, in
Section~\ref{horn};

\item the result in the general case extends to arbitrary doxastic states and
possibly heterogeneous sequences of revisions, in Section~\ref{mixed}.

\end{itemize}

Section~\ref{conclusions} summarizes the results of the article and suggests
possible future directions of study.

\section{Definitions}
\label{definitions}

\draft

definition of the doxastic state

\enddraft

Many iterated belief revision operators hinge on the strength of beliefs, the
doxastic state~\cite{rott-09}. Their most common formalization is a connected
(total) preorder between propositional models: $I \leq J$ means that the
situation $I$ is more firmly believed than the situation $J$ in the doxastic
state $C$.

Complete ignorance, like when learning about a completely new topic, is the
flat doxastic state: all situations are equally believed plausible. The
doxastic state is changed repeatedly as information is acquired in the form of
propositional formulae.

\begin{definition}
\label{flat}

The {\em flat order} $\emptyset$ compares all models the same.

\end{definition}

All formulae are assumed to be propositional formulae over a finite alphabet. A
literal is a symbol or the negation of a symbol. Models are truth evaluations on
this alphabet, and are denoted by the set of literals it sets to true. For
example, $\{a,\neg b\}$ is the model setting $a$ to $\true$ and $b$ to
$\false$.

The order of the doxastic state $D$ is denoted $\leq_D$. Its associated strict
order is denoted $<_D$, the equivalent relation $\equiv_D$. They are all
relations over the set of models, written in infix notation like
{} $\{a,b\} \leq_D \{\neg a, b\}$.

\draft

definition of the lexicographic order

\enddraft

Lexicographic revision increases the strenght of belief in all situations that
are consistent with the acquired information over those inconsistent with
it~\cite{spoh-88,naya-94}.

\begin{definition}
\label{order-formula}

The order induced by a propositional formula $F$ is defined by $I \leq_F J$ if
either $I \models F$ or $J \not\models F$, where $I$ and $J$ are propositional
models.

\end{definition}

A sequence of lexicographic revisions is dominated by its last, which makes the
bulk of the ordering, separating its satisfying models from the others. The
previous matter only for ties.

\begin{definition}
\label{lexicographic}

The doxastic state produced by the lexicographic revisions
{} $S = [S_1,\ldots,S_m]$
on the flat order $\emptyset$ compares $I \leq_{\emptyset S} J$ if:

\begin{itemize}

\item either $S=[]$ or

\item
\begin{itemize}

\item[] $I \leq_{S_m} J$ and

\item[]
either $J \not\leq_{S_m} I$
	or
$I \leq_{\emptyset R} J$, where $R = [S_1,\ldots,S_{m-1}]$.

\end{itemize}

\end{itemize}

\end{definition}

\draft

$
I \leq_F J \mbox{ is }
I \models F \mbox{ or } J \not\models F
$

$
I \leq_{\emptyset S} J \mbox{ is }
S=[]
\mbox{ or }
(
	I \leq_{S_1} J
	\mbox{ and }
	(
		J \not\leq_{S_1} I
		\mbox{ or }
		I \leq_{\emptyset R} J
	)
)
$

\enddraft

As an example, the empty doxastic state $\emptyset$ on the alphabet $\{a,b\}$
is changed by the lexicographic revision $a$ into $\emptyset[a]$, which
compares the propositional models $\{a,b\}$ and $\{a, \neg b\}$ strictly less
than all others, maintaining the remaining equal comparisons.

\begin{eqnarray*}
\{     a,     b\}	& \equiv_{\emptyset[a]} &	\{     a, \neg b\} \\
\{     a,\neg b\}	& <_{\emptyset[a]} &		\{\neg a,      b\} \\
\{\neg a,     b\}	& \equiv_{\emptyset[a]} &	\{\neg a, \neg b\}
\end{eqnarray*}

A further revision by $\neg b$ changes this ordering by making all models
setting $b$ to false strictly less than all others. The other comparisons stay
the same.

\begin{eqnarray*}
\{     a,\neg b\}	& <_{\emptyset[a, \neg b]} &	\{\neg a, \neg b\} \\
\{\neg a,\neg b\}	& <_{\emptyset[a, \neg b]} &	\{     a,      b\} \\
\{     a,     b\}	& <_{\emptyset[a, \neg b]} &	\{\neg a,      b\}
\end{eqnarray*}

\draft

definition of equivalence between sequences

\enddraft

The same order can result from different sequences of revisions. As an example,
{} $[\neg a \wedge \neg b, \neg a \wedge b, a \wedge \neg b, a \wedge b]$
produces the same order
{} $[a, b]$
does.

\begin{definition}[Equivalence]
\label{equivalence}

The sequences of lexicographic revisions $S$ and $R$ are equivalent, denoted $S
\equiv R$, if they produce the same order from the flat order: $I
\leq_{\emptyset S} J$ and $I \leq_{\emptyset R} J$ coincide for all pairs of
models $I$ and $J$.

\end{definition}

Sometimes, non-equivalence is easier to handle than equivalence: two sequences
$S$ and $R$ are not equivalent if $I \leq_S J$ and $I \not\leq_R J$ or the same
with $S$ and $R$ swapped for some models $I$ and $J$.

Redundancy of a revision in a sequence of lexicographic revisions is the
equivalence of the sequence with the result of removing the revision from it.

\section{Algorithm for equivalence}
\label{algorithm}

Solving problems by reduction to propositional satisfiability is a popular
strategy: the problem is converted into propositional satisfiability, which is
then solved by a complete satisfiability checker~\cite{darw-pipa-21}.

The translation solves the problems of equivalence and redundancy of two
sequences of lexicographic revision from the flat doxastic state. It indirectly
solves these problems on arbitrary doxastic states and possibly heterogeneous
sequences of revisions among natural, severe, moderate severe, plain severe,
restrained, very radical, full meet revision~\cite{rott-09,libe-23}.

\subsection{Summary}

The equivalence between sequences of lexicographic revisions is encoded by a
propositional formula. Each of its models is a counterexample to equivalence
between the orders they produce from the flat doxastic state: it contains two
models of the original alphabet that are sorted differently by the two orders.

The sequences are assumed of the same length for simplicity. This is not a
limitation since every sequence of lexicographic revisions can be elongated by
replicating its last formula.

The original alphabet is denoted $X$. Two disjoint copies of it are denoted $Y$
and $Z$. They allow the formula to express conditions over two models over $X$
at the same time. The formula is satisfied by a model over $Y \cup Z$ exactly
when they are not equivalent.

For every formula $F$ and sets of variables $X$ and $Y$ in one-to-one
correspondence, $F[X/Y]$ denotes the formula obtained by uniformly replacing
each variable in $Y$ with its corresponding variable in $X$.

\subsection{Model translation}

Every pair of models over $X$ translates into a single model over $Y \cup Z$
and back: if $X = \{x_1,\ldots,x_n\}$, then $Y = \{y_1,\ldots,y_n\}$ and $Z =
\{z_1,\ldots,z_n\}$. A model over $X$ assigns a truth value to every $x_i$; the
corresponding model over $Y$ assigns the same value to $y_i$. The same for $Z$.

The translation in one direction is just a replacement of variables. The
opposite conversion selects half of the variables: a model over $X$ is made of
the values on $Y$ only, discarding those of $Z$; another model is made of the
values of $Z$ only, discarding those of $Y$. This cutting is formalized by a
notation evoking substitution, but on models instead of formulae.

\begin{eqnarray*}
I[X/Y] &=&
	\{x_i = \true  \mid (y_i = \true)  \in I, y_i \in Y\} \cup 
	\{x_i = \false \mid (y_i = \false) \in I, y_i \in Y\} \\
\end{eqnarray*}

The resulting model $I[X/Y]$ assigns value to the variables $X$ only, and it
does according to the values of $Y$ in $I$. This notation includes both cutting
out the other variables and converting the values of $Y$ into the values of
$X$.

This notation is used in both directions:

\begin{description}

\def\temp{$I[Y/X]$}
\item[\temp] translates a model over the original alphabet $X$ into the one
over the first of the new ones $Y$;

\def\temp{$I[X/Y]$}
\item[\temp] translates a model over the new alphabets $Y \cup Z$ and possibly
other variables into a model over the original alphabet $X$ only.

\end{description}

The same do $I[Z/X]$ and $I[X/Z]$.

\subsection{Translation}

The translation turns every formula $S_i$ of the first sequence of
lexicographic revisions into $S_i[Y/X]$. If $I$ is a model of $S_i$ then
$I[Y/X]$ is a model of $S_i[Y/X]$. In the other way around, if $I'$ is a model
of $S_i[Y/X]$ then $I'[X/Y]$ is a model of $S_i$. The same goes for the second
sequence.

\begin{center}
\setlength{\unitlength}{2500sp}%
\begin{picture}(1230,861)(3661,-3139)
\thinlines
{\color[rgb]{0,0,0}\put(3751,-2686){\vector( 1, 0){1050}}
}%
{\color[rgb]{0,0,0}\put(4801,-2986){\vector(-1, 0){1050}}
}%
\put(4876,-2461){\makebox(0,0)[lb]{\smash{\fontsize{6}{7.2}
\usefont{T1}{cmr}{m}{n}{\color[rgb]{0,0,0}$S_i[Y/X]$}%
}}}
\put(4876,-3061){\makebox(0,0)[lb]{\smash{\fontsize{6}{7.2}
\usefont{T1}{cmr}{m}{n}{\color[rgb]{0,0,0}$I'$}%
}}}
\put(4876,-2761){\makebox(0,0)[lb]{\smash{\fontsize{6}{7.2}
\usefont{T1}{cmr}{m}{n}{\color[rgb]{0,0,0}$I[X/Y]$}%
}}}
\put(3676,-2461){\makebox(0,0)[rb]{\smash{\fontsize{6}{7.2}
\usefont{T1}{cmr}{m}{n}{\color[rgb]{0,0,0}$S_i$}%
}}}
\put(3676,-2761){\makebox(0,0)[rb]{\smash{\fontsize{6}{7.2}
\usefont{T1}{cmr}{m}{n}{\color[rgb]{0,0,0}$I$}%
}}}
\put(3676,-3061){\makebox(0,0)[rb]{\smash{\fontsize{6}{7.2}
\usefont{T1}{cmr}{m}{n}{\color[rgb]{0,0,0}$I'[X/Y]$}%
}}}
\end{picture}%
\nop{
   Si               Si[Y/X]
   I         --->   I[Y/X]
   I'[X/Y]   <---   I'
}
\end{center}

This correspondence is proved by the following lemma.

\state{rewriting}{lemma}{}{

\begin{lemma}
\statelabel

For every formula $F$ over variables $X$, every set of variables $Y$ of the
same cardinality of $X$ and every model $I$ that evaluates all variables in
$Y$, the following equality holds.

\[
I \models F[Y/X]	\mbox{ iff }	I[X/Y] \models F
\]

\end{lemma}

}{

\proof Since $X$ is the alphabet of $F$, the alphabet of $F[Y/X]$ is $Y$. This
formula contains only variables $Y$. As a result, $I \models F[Y/X]$ is the
same as $I' \models F[Y/X]$ where $I'$ is the model comprising the assignments
to the variables $Y$ in the model $I$.

Uniformly changing the names of variables has no semantical effect. Replacing
$x$ with $y$ everywhere in an assignment and a formula does not change
satisfaction. In this case, $I' \models F[Y/X]$ is the same as $I'' \models
(F[Y/X])[X/Y]$, where $I''$ and is $I'$ translated from $Y$ to $X$. The
translation $X/Y$ is the opposite of $Y/X$. Therefore, $(F[Y/X])[X/Y]$ is the
same as $F$. Since $I''$ is obtained from $I'$ by converting the variables $Y$
into $X$, and $I'$ is obtained from $I$ by cutting out all variables but $X$,
this model $I''$ is $I[X/Y]$.

As a result, the proven entailment $I'' \models (F[Y/X])[X/Y]$ is the same as
the claim $I[X/Y] \models F$.~\qed

}

The foundation of the lexicographic revision is the order induced by a single
formula, $I \leq_{S_i} J$ being defined as $I \models S_i$ or $J \not\models
S_i$. The lemma allows rewriting these two conditions on the alphabet $X$ as
single formula on $Y \cup Z$. The first $I \models S_i$ translates into
$S_i[Y/X]$; the second $J \not\models S_i$ translates into $\neg S_i[Z/X]$. The
whole formula is $S_i[Y/X] \vee \neg S_i[Z/X]$. This formula is satisfied by a
model $I'$ if and only if $I'[X/Y] \leq_{S_i} I'[X/Z]$.

This correspondence is not proved for $\leq_{S_i}$ but for its associated
strict order and equivalence relation:

\begin{eqnarray*}
STRICT(S_i,X,Y,Z)	&=&	S_i[Y/X] \wedge \neg S_i[Z/X] \\
EQUIV(S_i,X,Y,Z)	&=&	S_i[Y/X] \equiv S_i[Z/X]
\end{eqnarray*}

The first formula $STRICT(S_i,X,Y,Z)$ translates the definition of $<_{S_i}$
from $X$ into a single formula over $Y \cup Z$. The second $EQUIV(S_i,X,Y,Z)$
does the same for $\equiv_{S_i}$.

\state{strict-equiv-formula}{lemma}{}{

\begin{lemma}
\statelabel

The following holds for every formula $F$ over alphabet $X$ and every model $I$
over the alphabet $Y \cup Z$.

\begin{eqnarray*}
I \models STRICT(S_i,X,Y,Z)	\mbox{ iff }	I[X/Y] <_{S_i} I[X/Z]	\\
I \models EQUIV(S_i,X,Y,Z)	\mbox{ iff }	I[X/Y] \equiv_{S_i} I[X/Z]
\end{eqnarray*}

\end{lemma}

}{

\proof A model $I$ satisfies $STRICT(S_i,X,Y,Z)$ if and only if it satisfies
both $S_i[Y/X]$ and $\neg S_i[Z/X]$. Lemma~\ref{rewriting} proves $I \models
S_i[Y/X]$ the same as $I[X/Y] \models S_i$, and $I \models S_i[Z/X]$ the same
as $I[X/Z] \models S_i$. These two conditions are equivalent to $I[X/Y] <_{S_i}
I[X/Z]$.

The same argument proves that $I \models EQUIV(S_i,X,Y,Z)$ is the same as $I
\models S_i[Y/X]$ being equivalent to $I[X/Z] \models S_i$, which is the same
as $I[X/Y] \equiv_{S_i} I[X/Z]$.~\qed

}

The order $I \leq_{\emptyset S} J$ is equivalently redefined in terms of its
associated strict order and equivalence relation induced by its formulae. These
respectively translate into $STRICT(S_i,X,Y,Z)$ and $EQUIV(S_i,X,Y,Z)$.

\state{strict-equivalent}{lemma}{}{

\begin{lemma}
\statelabel

For every sequence of lexicographic revisions $S = [S_1,\ldots,S_m]$, the
condition $I \leq_{\emptyset S} J$ holds if and only if either
{}	$S=[]$,
{}	$I <_{S_m} J$ or
{}	both $I \equiv_{S_m} J$ and $I \leq_{\emptyset R} J$
hold, where $R = [S_1,\ldots,S_{m-1}]$ and $\equiv_{\emptyset S}$ and
$<_{\emptyset S}$ are respectively the equivalence relation and the strict
order associated with $\leq_{\emptyset S}$.

\end{lemma}

}{

\proof

The condition in the statement of the lemma is rewritten until it
becomes the definition of $I \leq_{\emptyset S} J$.

\begin{eqnarray*}
&&
S=[]
\mbox{ or }
I <_{S_m} J
\mbox{ or }
(
I \equiv_{S_m} J
\mbox{ and }
I \leq_{\emptyset R} J
)
\\
&&
S=[]
\mbox{ or }
(I \leq_{S_m} J \mbox{ and } J \not\leq_{S_m} I)
\mbox{ or }
(
(I \leq_{S_m} J \mbox{ and } J \leq_{S_m} I)
\mbox{ and }
I \leq_{\emptyset R} J
)
\\
&&
S=[]
\mbox{ or }
J \not\leq_{S_m} I
\mbox{ or }
(
(I \leq_{S_m} J \mbox{ and } J \leq_{S_m} I)
\mbox{ and }
I \leq_{\emptyset R} J
)
\\
&&
S=[]
\mbox{ or }
J \not\leq_{S_m} I
\mbox{ or }
(
I \leq_{S_m} J
\mbox{ and }
J \leq_{S_m} I
\mbox{ and }
I \leq_{\emptyset R} J
)
\\
&&
S=[]
\mbox{ or }
(
(
J \not\leq_{S_m} I
\mbox{ or }
I \leq_{S_m} J
)
\mbox{ and }
(
J \not\leq_{S_m} I
\mbox{ or }
J \leq_{S_m} I
)
\mbox{ and }
(
J \not\leq_{S_m} I
\mbox{ or }
I \leq_{\emptyset R} J
)
)
\\
&&
S=[]
\mbox{ or }
(
(
J \not\leq_{S_m} I
\mbox{ or }
I \leq_{S_m} J
)
\mbox{ and }
(
\true
)
\mbox{ and }
(
J \not\leq_{S_m} I
\mbox{ or }
I \leq_{\emptyset R} J
)
)
\\
&&
S=[]
\mbox{ or }
(
(
J \not\leq_{S_m} I
\mbox{ or }
I \leq_{S_m} J
)
\mbox{ and }
(
J \not\leq_{S_m} I
\mbox{ or }
I \leq_{\emptyset R} J
)
)
\\
&&
S=[]
\mbox{ or }
(
I \leq_{S_m} J
\mbox{ and }
(
J \not\leq_{S_m} I
\mbox{ or }
I \leq_{\emptyset R} J
)
)
\\
&&
\end{eqnarray*}

The last condition is the same as in the definition of the lexicographic
revision.

A couple of steps require some explanation.

Since the preorder is connected, if $J$ is not less than or equal to $I$ then
$I$ is less than or equal to $J$. In formulae, $J \not\leq_{\emptyset S} I$
implies $I \leq_{\emptyset S} J$.

The second step exploits this implication: ``$I \leq_{\emptyset S} J$ and $J
\not\leq_{\emptyset S} I$'' requires two conditions to be true at the same
time, but the second entails the first; the second not only is necessary
because it is part of the conjunction, but also sufficient as entails the
first.

The last step is similar, on a disjunction: $J \not\leq_{\emptyset S} I$ or $I
\leq_{\emptyset S} J$ requires either of two conditions to be true; the second
is not only sufficient because it is one of the two alternative cases; it is
also necessary because it is true whenever the first of the two alternatives is
the case.~\qed

}

This lemma redefines $\leq_{\emptyset S}$ in terms of $<_{\emptyset S_i}$ and
$\equiv_{\emptyset S_i}$, which respectively translate into the formulae%
{} $STRICT(S_i,X,Y,Z)$ and $EQUIV(S_i,X,Y,Z)$.
Combining these two steps gives a formula expressing $\leq_{\emptyset S}$ on
the alphabet $Y \cup Z$.

The starting point is storing the values of%
{} $STRICT(S_i,X,Y,Z)$ and $EQUIV(S_i,X,Y,Z)$
into new variables $m_i$ and $e_i$, which stand for ``minor'' and
``equivalent''. The order is then given by the following formula.

\[
ORDER(M,E) =
m_m \vee
(e_m \wedge
	(m_{m-1} \vee
	(e_{m-1} \wedge
		\cdots
	)
	)
)
\]

This formula is satisfied by a model $I$ over $Y \cup Z$ exactly when $I[X/Y]
\leq_S I[X/Z]$, if $m_i$ and $e_i$ have respectively the same values of
$STRICT(S_i,X,Y,Z)$ and $EQUIV(S_i,X,Y,Z)$.

\state{sequence-order-formula}{lemma}{}{

\begin{lemma}
\statelabel

A model $I$ that satisfies
{} $m_i \equiv STRICT(S_i,X,Y,Z)$
and
{} $e_i \equiv EQUIV(S_i,X,Y,Z)$
for all $i = 1,\ldots,m$
satisfies $ORDER(M,E)$ if and only if
{} $I[X/Y] \leq_S I[X/Z]$.

\end{lemma}

}{

\proof Lemma~\ref{strict-equiv-formula} proves that
{} $I \models STRICT(S_i,X,Y,Z)$
is the same as
{} $I[X/Y] <_{S_i} I[X/Z]$.
Therefore,
{} $I \models m_i \equiv STRICT(S_i,X,Y,Z)$
holds exactly when the truth value of $m_i$ in $I$ is the same as the condition
{} $I[X/Y] <_{S_i} I[X/Z]$
being the case.

For the same reason,
{} $I \models m_i \equiv EQUIV(S_i,X,Y,Z)$
holds exactly when the truth value of $m_i$ in $I$ is the same as the condition
{} $I[X/Y] \equiv_{S_i} I[X/Z]$
being the case.

Lemma~\ref{strict-equivalent} expresses $I[X/Y] \leq_S I[X/Z]$ as a combination
of conditions
{} $I[X/Y] <_{S_i} I[X/Z]$
and
{} $I[X/Y] \equiv_{S_i} I[X/Z]$.
Rewriting the former as $m_i$ and the latter as $e_i$ gives $ORDER(M, E)$.~\qed

}

This lemma translates $I[X/Y] \leq_S I[X/Z]$ into a boolean formula. The final
destination is a formula expressing the non-equivalence between two orders. It
is achieved by replicating the same construction for both orders and comparing
the result.

\begin{eqnarray*}
DIFF(S,R) &=&
\\
&& \bigwedge \{m_i \equiv STRICT(S_i,X,Y,Z) \mid i = 1,\ldots,m\}
\wedge\\
&& \bigwedge \{e_i \equiv EQUIV(S_i,X,Y,Z) \mid i = 1,\ldots,m\}
\wedge\\
&& (a \equiv ORDER(M,E))
\wedge\\
&& \bigwedge \{n_i \equiv STRICT(R_i,X,Y,Z) \mid i = 1,\ldots,m\}
\wedge\\
&& \bigwedge \{f_i \equiv EQUIV(R_i,X,Y,Z) \mid i = 1,\ldots,m\}
\wedge\\
&& (b \equiv ORDER(N,F))
\wedge\\
&& (a \not\equiv b)
\end{eqnarray*}

The correctness of $DIFF(S,R)$ is proved by the following lemma.

\state{difference-formula}{lemma}{}{

\begin{lemma}
\statelabel

A model $I$ satisfies $DIFF(S,R)$ if and only if the orders
{} $\leq_{\emptyset S}$ and $\leq_{\emptyset R}$
compare $I[X/Y]$ and $J[X/Z]$ differently.

\end{lemma}

}{

\proof Lemma~\ref{sequence-order-formula} proves that $I$ entailing
{} $m_i \equiv STRICT(S_i,X,Y,Z)$ and
{} $e_i \equiv EQUIV(S_i,X,Y,Z)$
for all $i = 1,\ldots,m$ make $I \models ORDER(M,E)$ the same as
{} $I[X/Y] \leq_S I[X/Z]$.
Therefore,
{} $I \models a \equiv ORDER(M,E)$
make $I \models a$ be the same as
{} $I[X/Y] \leq_S I[X/Z]$.

Similarly, $I$ satisfying
{} $n_i \equiv STRICT(R_i,X,Y,Z)$ and
{} $f_i \equiv EQUIV(R_i,X,Y,Z)$
for all $i = 1,\ldots,m$ and
{} $b \equiv ORDER(N,F)$
make $I \models b$ be the same as
{} $I[X/Y] \leq_R I[X/Z]$.

The final formula $a \not\equiv b$ is satisfied by $I$ if and only if
{} $\leq_{\emptyset S}$ and $\leq_{\emptyset R}$
compare $I[X/Y]$ and $I[X/Z]$ differently.~\qed

}

The model $I$ is an arbitrary assignment over the alphabet of $DIFF(S,R)$, which
comprises $Y$ and $Z$. As a result, $I[X/Y]$ and $I[X/Z]$ are two arbitrary
models over $X$. As a result, $DIFF(S,R)$ is satisfiable if and only if $I
\leq_S J$ differs from $I \leq_R J$ for some arbitrary pair of models $I$ and
$J$ over $X$. This completes the proof of correctness of the translation.

\state{equivalent-unsatisfiable}{theorem}{}{

\begin{theorem}
\statelabel

The sequences of lexicographic revisions $S$ and $R$ are equivalent if and only
if $DIFF(S,R)$ is unsatisfiable.

\end{theorem}

}{

\proof Lemma~\ref{difference-formula} proves that a model $I$ satisfies
$DIFF(S,R)$ if and only if $S$ and $R$ compare $I[X/Y]$ and $I[X/Z]$. Since $I$
is an arbitrary model, so are $I[X/Y]$ and $I[X/Z]$. As a result, a model
satisfies $DIFF(S,R)$ if and only if $S$ and $R$ compare two models
differently. In reverse, $DIFF(S,R)$ is unsatisfiable if and only if $S$ and
$R$ compare every pair of models the same. The latter is definition of
equivalence between sequences of revisions.~\qed

}

\subsection{Heterogeneous sequences of revisions from arbitrary orders}

The algorithm checks the equivalence of two sequences of lexicographic
revisions from the flat order. Every connected preorder is the result of an
appropriate sequence of lexicographic revisions on the flat
order~\cite{libe-24}. Heterogenous sequences of natural, very radical, severe,
moderate severe, plain severe, restrained and full meet revisions translate
into lexicographic~\cite{libe-23}.

The overall algorithm for checking the equivalence of $S$ and $R$ from the
doxastic state $C$ is:

\begin{itemize}

\item find a sequence $P$ of lexicographic revisions that produces $C$ from the
flat doxastic state~\cite{libe-23};

\item translate the concatenation of $P$ and $S$ into a sequence of
lexicographic revisions $L$~\cite{libe-23};

\item translate the concatenation of $P$ and $R$ into a sequence of
lexicographic revisions $L'$~\cite{libe-23};

\item translate $L$ and $L'$ into propositional unsatisfiability by the above
translation;

\item solve the unsatisfiability problem by one of the many existing and
efficient SAT solvers~\cite{darw-pipa-21}.

\end{itemize}

This algorithm establishes for example the redundancy of the third, natural
revision in the sequence comprising a first restrained revision, a second and a
third natural revisions and two lexicographic revisions followed by a very
radical revision from an arbitrary doxastic state. It tells whether forgetting
the third revision episode is harmless as it leads to no information loss. Or
useless, if information loss was the aim of forgetting rather than a drawback.

\section{A necessary and sufficient condition to redundancy}
\label{condition}

The irredundancy of the first of a sequence of lexicographic revision is
expressed by the existence of two models comparing different by the first
revision and the same by the others.

\state{relevance}{theorem}{}{

\begin{theorem}
\statelabel

The first formula of $S=[S_1,S_2,\ldots,S_m]$ is irredundant if and only if
{} $I \not\equiv_{S_1} J$
and
{} $I \equiv_{S_i} J$ where $i = 2,\ldots,m$
for some models $I$ and $J$.

\end{theorem}

}{

\proof

When two such models exists, the definition of $I \leq_{\emptyset S} J$
specializes as follows:

\begin{eqnarray*}
\lefteqn{I \leq_{\emptyset S} J}
\\
&\mbox{iff}&
I \leq_{\emptyset S_m} J \mbox{ and either }
J \not\leq_{\emptyset S_m} \mbox{ or }
I \leq_{\emptyset [S_1,\ldots,S_{m-1}]} J
\\
&\mbox{iff}&
\true \mbox{ and either }
\false \mbox{ or } I \leq_{\emptyset [S_1,\ldots,S_{m-1}]} J
\\
&\mbox{iff}&
I \leq_{\emptyset [S_1,\ldots,S_{m-1}]} J
\end{eqnarray*}

Inductively, the same holds for $I \leq_{\emptyset [S_1,\ldots,S_i]} J$
when $i=m,\ldots,2$. The outcome is $I \leq_{\emptyset S_1} J$. Therefore, $I
\leq_{\emptyset S} J$ is the same as $I \leq_{\emptyset S_1} J$.

The same deduction applies to $J \leq_{\emptyset S} I$, which is equivalent to
$J \leq_{\emptyset S_1} I$. The conclusion is that $I \equiv_{\emptyset S} J$
is the same as $I \equiv_{S_1} J$, which is false.

The same deduction equates $I \equiv_{\emptyset [S_2,\ldots,S_m]} J$ with $I
\equiv_{\emptyset S_2} J$, which is true. The conclusion is that $I$ and $J$
are equivalent according to $\leq_{\emptyset [S_2,\ldots,S_m]}$ but not to
$\leq_{\emptyset [S_1,\ldots,S_m]}$: the first formula of $S$ is not redundant.

\

If no two such model exists, then $I \not\equiv{\emptyset S_i} J$ holds for
some $i=2,\ldots,m$ for all models $I$ and $J$. The highest such index $i$ is
considered: $I \equiv_{S_j} J$ for all $j=i+1,\ldots,m$. The assumption $I
\not\equiv{\emptyset S_i} J$ implies either
{} $I \not\leq_{\emptyset S_i} J$
or
{} $J \not\leq_{\emptyset S_i} I$.
Only the first case is assumed, the second mimics the first by symmetry.

Since $I \equiv_{\emptyset S_j} J$ holds for $j=i+1,\ldots,m$, the deduction
above equates $I \leq_{\emptyset S} J$ with $I \leq_{[S_1,\ldots,S_i]} J$,
which is false because of
{} $I \not\leq_{\emptyset S_i} J$.

For the same reason, $I \leq_{\emptyset [S_2,\ldots,S_m]} J$ is false.

This proves the equivalence of $S$ and $[S_2,\ldots,S_m]$: the first formula of
$S$ is redundant.

\qed

}

\section{Complexity}
\label{general}

Establishing the equivalence between two sequences of lexicographic revisions
is a \conp-complete problem. First proved is membership.

\state{general-membership}{lemma}{}{

\begin{lemma}
\statelabel

Checking whether two sequences of lexicographic revisions are equivalent is in
\conp.

\end{lemma}

}{

\proof The opposite to equivalence is the existence of two models $I$ and $J$
compared differently by the two orders: $I \leq_{\emptyset S} J$ and not $I
\leq_{\emptyset Q} J$.

Checking $I \leq_{\emptyset S} J$ takes polynomial time. It requires checking
$I \leq_{S_m} J$, $J \leq_{S_m} I$ and $I \leq_{\emptyset R} J$, where $S_m$ is
the last formula of $S$ and $R$ is the sequence of the preceding ones. Checking
$I \leq_{S_m} J$ amounts to verify $I \models S_m$ and $J \models S_m$, both
linear-time operations. The checks $I \leq_{S_m} J$ and $J \leq_{S_m} I$ are
therefore linear-time operations.

Checking $I \leq_{\emptyset S} J$ therefore takes linear time for the first
formula of $S$ and the time of the recursive call on the rest of the sequence.
Since a formula is removed at every call, the number of recursive calls is
linear. A linear number of calls, each requiring linear time. Overall, the
algorithm takes polynomial time.

Equivalence requires comparing $I \leq_{\emptyset S} J$ and $I \leq_{\emptyset
Q} J$ for all pairs of models $I$ and $J$. It amounts to check two orders for
all pairs of models. Since checking the order between two models takes only
polynomial time, the overall problem is in \conp.~\qed

}

Equivalence is also hard to \conp. Same for redundancy.

\state{general-hard}{lemma}{}{

\begin{lemma}
\statelabel

Checking the equivalence of two sequences of lexicographic revisions is
\conp-hard, and remains hard even when the second sequence is the first without
an element.

\end{lemma}

}{

\proof The claim is proved by reduction from the problem of validity of a
propositional formula $F$. Namely, the validity of $F$ is the same as the
equivalence of $[F_1,F_2]$ with $[F_2]$, where the formulae of the sequences
are as follows and $x$ and $y$ are fresh variables.

\begin{eqnarray*}
F_1 &=& x \vee (\neg x \wedge y \wedge F) \\
F_2 &=& x
\end{eqnarray*}

Every unsatisfiable formula $F$ is equivalent to $\false$. Therefore,
{} $F_1 = x \vee (\neg x \wedge y \wedge F)$
is equivalent to
{} $x \vee (\neg x \wedge y \wedge \false)$,
which is equivalent to
{} $x \vee \false$
and to $x$. The order $\leq_{\emptyset [F_1,F_2]}$ is $\leq_{\emptyset [x,x]}$,
which coincides with $\leq_{\emptyset [x]}$, which is $\leq_{\emptyset [F_2]}$.


If $F$ is satisfiable, it has a model. It extension by $x=\false$ and $y=\true$
is denoted $I$. A model setting $x=\false$, $y=\true$ and the remaining
variables to false is denoted $J$.

These two models meet the premises of Theorem~\ref{relevance}: since they both
falsify $y$, they are equivalent according to $\leq_{\emptyset F_2}$. Since $I$
satisfy $y$ and $J$ falsifies it, they are not equivalent according to
$\leq_{\emptyset F_1}$. Theorem~\ref{relevance} proves the irredundancy of
$F_1$ in $\leq_{\emptyset [F_1,F_2]}$.

\qed

}

Membership and hardness define completeness.

\begin{corollary}
\label{general-redundancy-complete}

Checking whether a revision is redundant in a sequence of lexicographic
revisions is \conp-complete.

\end{corollary}

\begin{corollary}
\label{general-equivalence-complete}

Checking whether two sequences of lexicographic revisions are equivalent is
\conp-complete.

\end{corollary}

\section{Horn sequences}
\label{horn}

Equivalence and redundancy of sequences of Horn lexicographic revisions is
\conp-complete, like for arbitrary propositional formulae.

A literal is negative if it is the negation of a variable, otherwise it is
positive. A clause is a disjunction of literals. It is positive if it comprises
only positive literals. It is Horn if it comprises no more than one positive
literal. A Horn formula is a conjunction of Horn clauses, denoted by their set.

The problem of checking the redundancy of Horn lexicographic revision belongs
to \conp\  because it does in the general case, as proved by
Lemma~\ref{general-membership}. Hardness is proved by the following lemma.

\state{horn-hard}{lemma}{}{

\begin{lemma}
\statelabel

Checking whether a revision is redundant in a sequence of lexicographic Horn
revisions is \conp-hard.

\end{lemma}

}{

\proof The claim is proved by reduction from propositional unsatisfiability: a
propositional CNF formula $F$ is unsatisfiable if and only if
{} $[S_1,S_2,\ldots,S_m] \equiv [S_2,\ldots,S_m]$
for some formulae $S_1,\ldots,S_m$.

These formulae are generated from $F$ by replacing each positive literal $x_i$
with the negation $\neg x_i'$ of a fresh variable and adding the clause
{} $\neg x_i \vee \neg x_i'$.
The result is denoted $F^n$. It is Horn since it contains only negative
literals.

\[
F^n =
	\{c[\neg x_i'/x_i] \mid c \in F\}
	\cup
	\{\neg x_i \vee \neg x_i' \mid x_i \in X\}
\]

The disjunction $\neg y \vee F$ of a literal with a formula is
{} $\{\neg y \vee c \mid c \in F\}$.
It is Horn if $F$ is Horn.

The reduction is the following.

\begin{eqnarray*}
S_1	&=&	y					\\
S_2	&=&	y \wedge \neg x_1 \wedge \neg x_1'	\\
	&\vdots&					\\
S_{m-1}	&=&	y \wedge \neg x_n \wedge \neg x_n'	\\
S_m	&=&	\neg y \vee F^n
\end{eqnarray*}

Theorem~\ref{relevance} establishes the irredundancy of $S_1$ to be the same as
the existence of two models $I$ and $J$ such that $I \equiv_{\emptyset S_i} J$
for all $i = 2,\ldots,m$ and
{} $I \not\equiv_{\emptyset S_1} J$.

The latter condition
{} $I \not\equiv_{\emptyset S_1} J$
implies that either $I \models y$ and $J \not\models y$ or the other way
around. Symmetry allows neglecting the second case.

Since $J$ falsifies $y$, it also falsifies $S_i = y \wedge \neg x_i \wedge \neg
x_i'$ for all $i=2,\ldots,m-1$. Since $I \equiv_{S_i} J$, the same applies to
$I$: it falsifies $y \wedge \neg x_i \wedge \neg x_i'$. Since it satisfies $y$,
it falsifies $\neg x_i \wedge \neg x_i'$. It therefore satisfies $x_i \vee
x_i'$.

Since $J$ falsifies $y$, it satisfies $S_m = \neg y \vee F^n$. Because of
$I \equiv_{S_m} J$, also $I$ satisfies $\neg y \vee F^n$. Since it falsifies
$\neg y$, it satisfies $F^n$.

The conclusion is that $I$ satisfies $F^n$ and all formulae $x_i \vee x_i'$.
The conjunction of these formulae is denoted $Q$. The formula $F^n$ contains
all clauses $\neg x_i \vee \neg x_i'$. With the clauses $x_i \vee x_i'$, they
are equivalent to $x_i' \equiv \neg x_i$. As a result, $Q$ is equivalent to
{} $F \wedge \{x_i' \equiv \neg x_i \mid x_i \in X\}$.
This formula is satisfiable if and only if $F$ is satisfiable.

This equates the satisfiability of $F$ with the irredundancy of $S_1$ in
$[S_1,\ldots,S_m]$. Establishing the redundancy of an element of a sequence of
Horn formulae is \conp-hard.

\qed

}

Membership and hardness is completeness.

\state{horn-complete}{theorem}{}{

\begin{theorem}
\statelabel

Checking whether a revision is redundant in a sequence of lexicographic Horn
revisions is \conp-complete.

\end{theorem}

}{

\proof The problem is in \conp\  for arbitrary formulae as proved by
Lemma~\ref{general-membership}. Therefore, it is also in \conp\  for Horn
formulae. Lemma~\ref{horn-hard} proves it \conp-hard for Horn formulae.~\qed

}

\section{Homogeneous or heterogeneous sequences of revisions}
\label{mixed}

\draft
{\bf   motivations   }
\enddraft

Lexicographic revision is central to iterated belief revision not only for its
historical role but also for its wide applicability, as it formalize the
introduction of a belief pertaining to all situations and not only the
currently believed ones~\cite{libe-23-b}. Yet, other revisions exist. Natural
revision~\cite{bout-96-a} introduces beliefs pertaining only to the present
conditions, not to all. Radical and moderate revisions~\cite{rott-09} make yet
differing assumptions.

\draft
{\bf   membership    }
\enddraft

In spite of all their differences, they all reduce to lexicographic revisions
in a way or the other~\cite{libe-23}. Yet, the translation requires solving a
number of satisfiability problems. Establishing the equivalence of two possibly
heterogeneous sequences of a number of revisions amounts to translate both into
sequences of lexicographic revisions only. The resulting equivalence problem
requires only a further satisfiability test thanks to
Theorem~\ref{general-equivalence-complete}. Polynomiality with the aid of an
external satisfiability checker is a definition of the complexity class \D2.

\state{heterogeneous-equivalence-membership}{theorem}{}{

\begin{theorem}
\statelabel

Establishing the equivalence of two possibly heterogeneous sequences of
revisions among natural, very radical, severe, plain severe, moderate severe
and full meet is in \D2.

\end{theorem}

}{

\proof The problem of checking the equivalence of $S$ and $S'$ is solved by
first translating both sequences into two sequences of lexicographic revisions
$L$ and $L'$, which are then checked for equivalence.

\begin{itemize}

\item Translating $S$ into $L$ requires polynomial time except the time for
solving a polynomial number of problems in \np~\cite{libe-23}.

\item The same for translating $S'$ into $L'$.

\item Equivalence of $L$ and $L'$ is the contrary of non-equivalence, a problem
in \np\  by Theorem~\ref{general-equivalence-complete}.

\end{itemize}

Polynomial time suffice with the addition of solving a polynomial number of
subproblems. This is one of the possible definitions of \D2.~\qed

}

Redundancy is a subcase of equivalence, the case where one of the two sequence
is the same as the other except for the presence of one formula. It therefore
belongs to the same complexity class.

\begin{theorem}
\label{heterogeneous-redundancy-membership}

Establishing the redundancy of a homogeneous or heterogeneous sequence of
revisions among natural, very radical, severe, plain severe, moderate severe
and full meet is in \D2.

\end{theorem}

\draft
{\bf    hardness     }
\enddraft

Lexicographic and radical revisions distinguish from the others for the
polynomiality of comparing two models: $I \leq J$ depends only on whether $I$
and $J$ satisfy some propositional formulae. Other revisions like natural
revision hinge on the minimal models of the revisions. Still others also on its
maximal ones. Minimal and maximal models are not only a matter of a model
satisfying a formula or not, but also of the existence of arbitrary other
models that do. "Arbitrary models with a property" is almost the definition of
the \np\  and \conp\  complexity classes. Given that $I \leq J$ may not be in
\p, the problem of equivalence of two sequences of revisions cannot be expected
to be in \conp\  like for lexicographic revisions only.

The complexity analysis bounds its complexity between \conp\  and \D2.
Membership in \D2\  is established by the above
Theorem~\ref{heterogeneous-equivalence-membership}, hardness to \conp\  by
Theorem~\ref{heterogeneous-redundancy-conphard}. The dependency to minimality
checks suggests hardness to \D2, but the current analysis failed to prove it to
date.

While the hardness level established so far is disappointingly equal to that
for lexicographic revisions only, its proof extends it significantly by
showing \conp-hardness not only for the considered revisions, but for all
revisions having two simple properties. Not only natural, radical, moderate and
full meet revision, but any other revision possessing these properties make the
redundancy and equivalence problems \conp-hard.

The first property is that a second revision that is exactly the opposite of
first erases it starting from the flat doxastic state. This is expected for
most belief revisions: only a formula and its negation are ever mentioned.
There is no reason to believe anything else.

The second property is that consistent and differing revisions both counts.
None contradicts the other. There is no reason to neglect one.

\state{heterogeneous-redundancy-conphard-conditions}{lemma}{}{

\begin{lemma}
\statelabel

Establishing the redundancy of a sequence of possibly heterogeneous revisions
is \conp-hard if the revisions satisfy the conditions:

\begin{itemize}

\item $[A,\neg A]$ is equivalent to $[\neg A]$;

\item $[A,B]$ is not equivalent to $[B]$
if $A \wedge B$ is consistent and not equivalent to $B$.

\end{itemize}

\end{lemma}

}{

\proof

The sequence $[a,a \rightarrow F]$ is proved redundant if and only if $F$ is
unsatisfiable, where $a$ is a fresh variable not contained in $F$.

If $F$ is inconsistent then $a \rightarrow F$ is equivalent to $\neg a$. The
sequence of revisions is therefore $[a,\neg a]$. It is equivalent to $[\neg a]$
by the first assumption of the lemma: the first revision is redundant.

If $F$ is consistent then
{} $a \wedge (a \rightarrow F)$
is equivalent to
{} $a \wedge F$.
This formula is consistent because $a$ and $F$ both are and do not share
variables. It is not equivalent to $a \rightarrow F$ because it is falsified by
all models falsifying $a$ while $a \rightarrow F$ is satisfied by them. The
sequence
{} $[a, a \rightarrow F]$
is not equivalent to
{} $[a \rightarrow F]$
by the second assumption of the lemma: the first revision is not redundant.

\qed

}

The two required properties are satisfied by lexicographic, natural, radical,
severe and full meet revisions.

\begin{theorem}
\label{heterogeneous-redundancy-conphard}

Establishing the redundancy of a possibly heterogeneous sequence of revisions
among natural, very radical, severe, plain severe, moderate severe and full
meet is \conp-hard.

\end{theorem}

Redundancy is a subcase of equivalence. Hardness for the former extends to the
latter.

\begin{theorem}
\label{heterogeneous-equivalence-conphard}

Establishing the equivalence of two possibly heterogeneous sequences of
revisions among natural, very radical, severe, plain severe, moderate severe
and full meet is \conp-hard.

\end{theorem}

\section{Conclusions}
\label{conclusions}

\draft
{\bf SUMMARY OF ARTICLE}
\enddraft

Forgetting an information acquisition episode may or may not lead to
information loss. This may be the very aim of forgetting, an unintended
consequence or just unwanted. Checking when it does is what the algorithm in
Section~\ref{algorithm} does. The problem is established \conp-complete both
for arbitrary formulae in Section~\ref{general} and in the Horn case in
Section~\ref{horn}. The results extend from the flat initial doxastic state
and lexicographic revisions only to arbitrary initial states and possibly
heterogeneous sequences of lexicographic, natural, restrained, severe, plain
severe, moderate severe, very radical and full meet revisions in
Section~\ref{mixed}.

\draft
{\bf FUTURE DIRECTIONS} 
\enddraft

Many future directions of study exist. While the problem is proved \conp-hard
already for two revisions lexicographic in the general case, it is only proved
hard for a linear number of lexicographic revisions in the Horn case. The
complexity for short Horn is still open. Other restrictions are important, like
Krom~\cite{krom-67} and others in the Post lattice~\cite{post-41}. Complexity
for heterogeneous revisions has a large gap: it is \conp-hard and belongs to
\D2, but could be complete for any class in between. A proof of \np-hardness
would at least prove redundancy and equivalence harder than for lexicographic
revisions only.

Redundancy is a relevant subcase of equivalence, which is \conp-complete. It is
also a relevant subcase of reducibility: shortening a sequence while
maintaining the order. Its decision version is whether an equivalent sequence
is smaller than it, or it is shorter than a given number. This problem can be
seen as generalizing Boolean formulae minimization from sequences of single
formulae to sequences of many~%
\cite{mccl-56,rude-sang-87,coud-94,theo-etal-96,coud-sasa-02,uman-etal-06}.
A further generalization is to allow some loss of doxastic information in
exchange for size reduction.

Redundancy matters also when the doxastic state is not a total preorder between
models~%
\cite{arav-etal-18,andr-etal-02,gura-kodi-18,souz-etal-19},
%
%
%
%
%
such as prioritized bases~%
\cite{brew-89,nebe-91,benf-etal-93},
conditionals~%
\cite{kuts-19,andr-etal-02,saue-etal-22}
and others coming from the preference reasoning field~%
\cite{doms-etal-11},
%
%
%
%
%
or abstract structures~%
\cite{schw-etal-22}.
%
%



\bibliographystyle{alpha}
\newcommand{\etalchar}[1]{$^{#1}$}

\end{document}